\documentclass[letterpaper, 10 pt, conference]{ieeeconf}
\usepackage{times}

\IEEEoverridecommandlockouts                              

\usepackage{multicol}
\usepackage[bookmarks=true]{hyperref}
\usepackage{xcolor}
\usepackage{subfigure}
\usepackage{graphicx}
\usepackage{amsmath}

\usepackage{algorithm}
\usepackage{algpseudocode}
\usepackage{caption}

\title{\LARGE \bf
Negotiation-based Human-Robot Collaboration via Augmented Reality
}

\author{Kishan Chandan$^{1}$, Vidisha Kudalkar$^{1}$, Xiang Li$^{2}$, Shiqi Zhang$^{1}$
\thanks{$^{1}$Department of Computer Science,
        SUNY Binghamton.
        Email: {\tt\small \{kchanda2, vkudalk1, zhangs\}@binghamton.edu}}%
\thanks{$^{2}$OPPO US Research Center. 
        Email: {\tt\small xiangli.sky@gmail.com}}%
}

\begin{document}

\maketitle
\thispagestyle{empty}
\pagestyle{empty}

\begin{abstract}

Effective human-robot collaboration (HRC) requires extensive communication among the human and robot teammates, because their actions can potentially produce conflicts, synergies, or both.
We develop a novel augmented reality (AR) interface to bridge the communication gap between human and robot teammates. 
Building on our AR interface, we develop an AR-mediated, negotiation-based (ARN) framework for HRC. 
We have conducted experiments both in simulation and on real robots in an office environment, where multiple mobile robots work on delivery tasks. 
The robots could not complete the tasks on their own, but sometimes need help from their human teammate, rendering human-robot collaboration necessary. 
Results suggest that ARN significantly reduced the human-robot team's task completion time compared to a non-AR baseline approach.

\end{abstract}

\IEEEpeerreviewmaketitle

\section{Introduction}

Robots are increasingly ubiquitous in everyday environments, but few of them collaborate or even communicate with people in their work time. 
For instance, Amazon's warehouse robots and people have completely separated work zones~\cite{wurman2008coordinating}. 
Savioke's Relay robots, having completed hundreds of thousands of deliveries in indoor environments, such as hotels and hospitals, do not interact with people until the moment of delivery~\cite{ivanov2017adoption}. 
Despite the significant achievements in multi-agent systems~\cite{wooldridge2009introduction}, human-robot collaboration (HRC), as a kind of multi-agent system, is still rare in practice.

Augmented Reality (AR) focuses on overlaying information in an augmented layer over the real environment to make the objects interactive~\cite{azuma2001recent}. 
On the one hand, AR has promising applications in robotics, and people can visualize the state of the robot in a visually enhanced form while giving feedback at runtime~\cite{green2007augmented}. 
On the other hand, there are a number of collaboration algorithms developed for multiagent systems (MAS)~\cite{wooldridge2009introduction,stone2000multiagent}, where a human-robot team is a kind of MAS. 
Despite the existing research on AR in robotics and multiagent systems, very few have leveraged AR for HRC (see Section~\ref{sec:related}). 
In this work, we develop an augmented reality-mediated, negotiation-based ({\bf ARN}) framework for HRC problems, where ARN for the first time enables spatially-distant, human-robot teammates to iteratively communicate preferences and constraints toward effective collaborations.

The AR interface of ARN enables the human teammate to visualize the robots' current status (e.g., their current locations) as well as the planned motion trajectories. 
For instance, \emph{a human user might ``see through'' a heavy door (via AR) and find a robot waiting for him/her to open the door.}
Moreover, ARN also supports people giving feedback to the robots' current plans. 
For instance, if the user is too busy to help on the door, he/she can indicate ``\emph{I cannot open the door for you in three minutes}'' using ARN. 
Accordingly, the robots will incorporate such human feedback for re-planning, and see if it makes sense to work on something else and come back after three minutes. 
The AR interface is particularly useful in environments with challenging visibility, such as the indoor environments of offices, warehouses, and homes, because the human might frequently find it impossible to directly observe the robots' status due to occlusions. 

ARN has been implemented both in simulation and with human participants, where we used a human-robot collaborative delivery task for evaluation.
Both human and robots are assigned non-transferable tasks.
Results from the real-world experiment suggest that ARN significantly improves the efficiency of human-robot collaboration, in comparison to traditional non-AR baselines. 
Additionally, we carried out experiments in simulation to evaluate two configurations of ARN (with and without human feedback). 
From the results, we observe the capability of taking human feedback significantly improved the efficiency of human-robot collaboration, in comparison to the ``no feedback'' configuration. 

\section{Related Work}
\label{sec:related}
When humans and robots work in a shared environment, it is vital that they communicate with each other to avoid conflicts, leverage complementary capabilities, and facilitate the smooth accomplishment of tasks. 
However, humans and robots prefer different modalities for communication. 
While humans employ natural language, body language, gestures, written communication, etc., the robots need information in a digital form, e.g., text-based commands. 
Researchers developed algorithms to bridge the human-robot communication gap using natural language ~\cite{chai2014collaborative,thomason2015learning,matuszek2013learning,amiri2019augmenting}
and vision ~\cite{Waldherr2000,nickel2007visual,yang2007gesture}. 
Despite those successes, AR has its unique advantages in elevating coordination through communicating spatial information, e.g., through which door a robot is coming into a room and how (i.e., the planned trajectory), when people and robots share a physical environment~\cite{azuma1997survey}. 
We use an AR interface for human-robot collaboration, where the human can directly visualize and interact with the robots' planned actions. 

One way of delivering spatial information related to the local environment is through projecting the robot's state and motion intent to the humans using visual cues~\cite{Park:2009:RPA:1514095.1514146,7354195,4373930}.
For instance, researchers used a LED projector attached to the robot to show its planned motion trajectory, allowing the human partner to respond to the robot's plan to avoid possible collisions~\cite{chadalavada2015s}. 
While such systems facilitate human-robot communication about spatial information, they have the requirement that the human must be in close proximity to the robot. 
Also, bidirectional communication is difficult in projection-based systems. 
We develop our AR-based framework that inherits the benefits of spatial information from the projection-based systems while alleviating the proximity requirement, and enabling bidirectional communication. 


Early research on AR-based human-robot interaction (HRI) has enabled a human operator to interactively plan, and optimize robot trajectories~\cite{583833}.
More recently, researchers have developed frameworks to help human operators to visualize the motion-level intentions of unmanned aerial vehicles (UAVs) using AR~\cite{walker2018communicating,hedayati2018improving}. 
In another line of research, people used an AR interface to help humans visualize a robot arm's planned actions in the car assembly tasks~\cite{amor2018intention}.
However, the communication of those systems is unidirectional, i.e., their methods only convey the robot's intention to the human but do not support the communication the other way around. 
Our ARN framework supports bidirectional communication toward effective collaborations.

Most relevant to this paper is a system that supports a human user to visualize the robot's sensory information, and planned trajectories, while allowing the robot to prompt information, as well as ask questions through an AR interface~\cite{muhammad2019creating,cheli2018}. 
In comparison to their work, our ARN framework supports human-multi-robot collaboration, where the robots collaborate with both robot and human teammates. 
More importantly, our robots are equipped with the task (re)planning capability, which enables the robots to respond to the human feedback by adjusting their task completion strategy.
Our robots' task planning capability enables negotiation and collaboration behaviors within human multi-robot teams.

In the next two sections, we present the major contributions of this research, including the AR interface for human-robot communication, and a human-robot collaboration framework based on the AR interface.

\section{Our AR Interface for Human-Robot Communication}
\label{sec:ar_interface}

\begin{figure}[tb]
\begin{center}
    \vspace{.8em}
    \includegraphics[width=.45\textwidth]{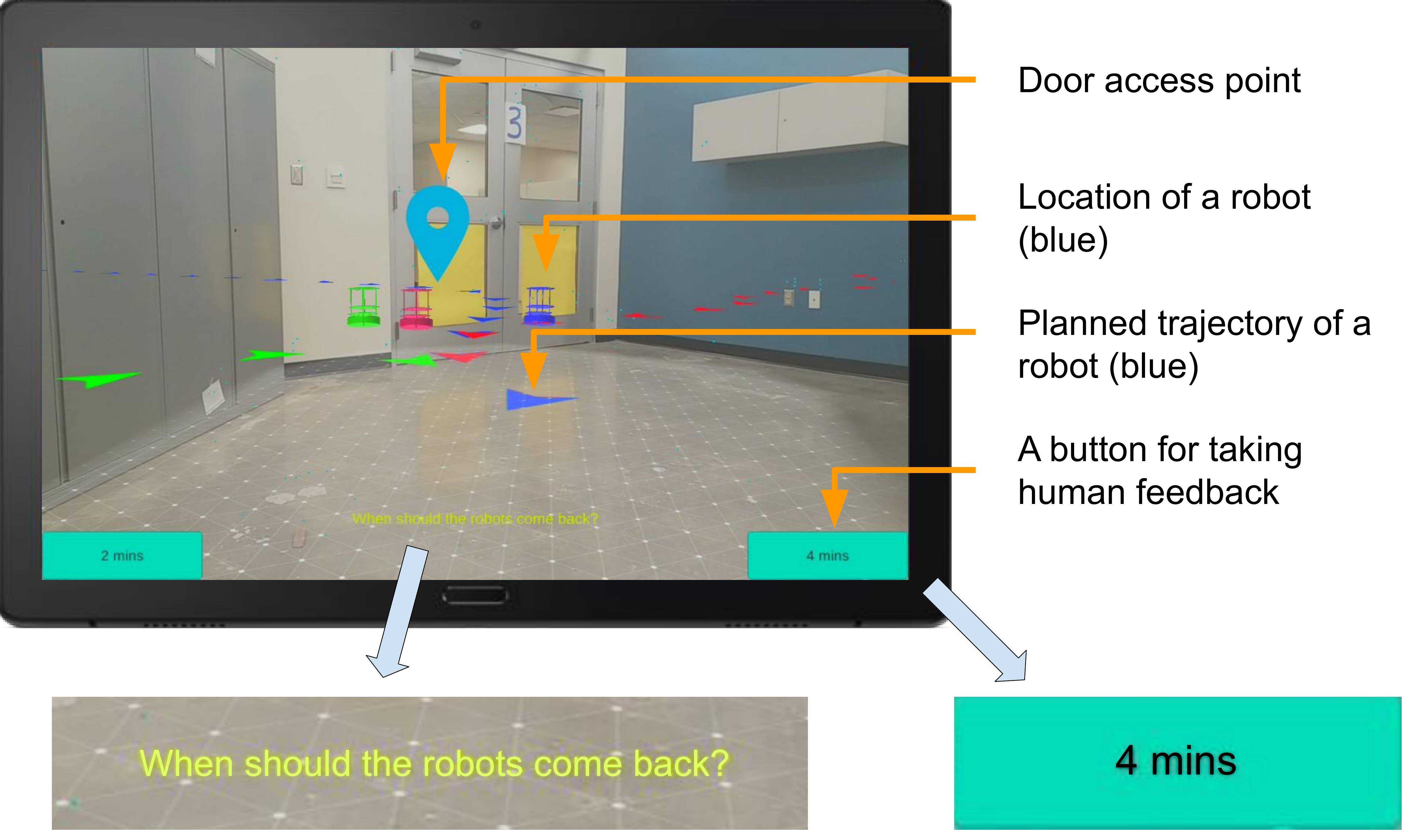}
    \vspace{-.5em}
    \caption{Our AR interface as displayed on a mobile device. The interface is used for visualizing the robots' locations and planned actions, and taking human feedback using the two interactive buttons at the bottom.}
    \label{fig:visualizer_figure}
    \end{center}
    \vspace{-1.5em}
\end{figure}

We have summarized a sample of existing research on human-robot communication, including their limitations. 
Toward less ambiguity and higher bandwidth, we develop a novel AR interface for human multi-robot communication, as shown in Fig.~\ref{fig:visualizer_figure}. 
Our AR interface, for the first time, enables the human teammate to visualize the robots' current state of the world, and their intended plans, while further allowing the human to give feedback on these plans. 
The robots can then use this feedback to adjust their task completion strategy, if necessary. Next, we focus on the two most important components of our AR interface, namely \emph{Visualizer} for visualizing robots' current states and intended plans, and \emph{Restrictor} for taking the human feedback for plan improvements. 

\vspace{.8em}
\noindent{\bf{Visualizer: }}
\label{sec:vslz}
The AR interface obtains a set of motion plans ($\boldsymbol{Trj}$), along with the live locations from the robots ($\boldsymbol{Pose}$).
The AR interface augments these motion plans as trajectories, and the live locations as visualizable 3D objects (robot avatars) with the $\boldsymbol{Vslz}$ function:
$$
V \xleftarrow{} \boldsymbol{Vslz}(\boldsymbol{Trj, Pose})
$$
At every time step $t$, these set of trajectories and robot avatars are rendered as $V$ on the AR interface, where $V$ is a single frame.

\begin{figure*}
\begin{center}
\vspace{0.5em}
    \includegraphics[width=14cm]{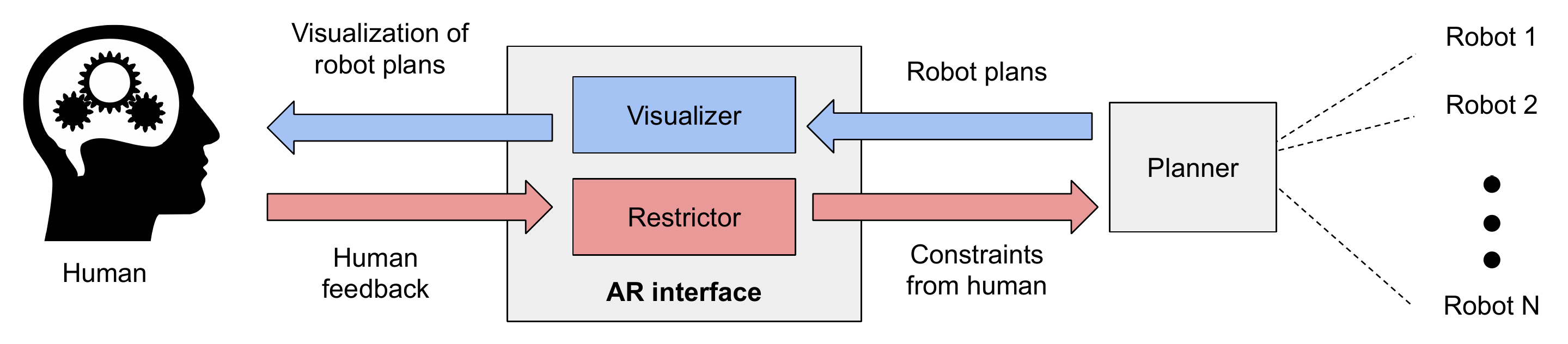}
    \vspace{-.5em}
\caption{Key components of our ARN framework: \emph{Visualizer} and \emph{Restrictor} for visualizing robot's intention (for people) and collecting human feedback (for robots) respectively, and \emph{Planner} for computing one action sequence for each robot. }
\label{fig:framework}
    \end{center}
    \vspace{-1.5em}
\end{figure*}

\vspace{.8em}
\noindent{\bf{Restrictor: }}
\label{sec:cnsts}
Once the robots' plans and live locations are augmented, the human can visualize them through the AR device.
The AR interface allows the human to also give feedback on the robots' plans.
These feedbacks are domain-specific and can involve speech, gestures, interactive buttons, etc. as modes of human input.
All the possible feedback is maintained in a feedback library ($F$).
The AR interface constantly checks for human feedback, which is then stored in $H$ using the $\boldsymbol{getFdk}$ function, where $H \in F$:
$$
H \xleftarrow{} \boldsymbol{getFdk}()
$$
If there is no human feedback, then $H$ is $\emptyset$.
Human feedback is communicated back to the robots, which can then be used for re-planning if necessary.
The human feedback in its original form cannot be used, and hence this human feedback needs to be converted to a set of activated constraints.
$$
C \xleftarrow{} \boldsymbol{Cnsts}(H)
$$
The $\boldsymbol{Cnsts}$ function takes the human feedback $H$ as input.
The output of $\boldsymbol{Cnsts}$ is stored in $C$, where $C$ is a set of activated constraints.
The new set of activated constraints can now be used by robots to re-plan if required.

The two components of our AR interface enable the visualization of the robots’ status and the incorporation of human feedback.
In the next section, we describe how we use the AR interface to develop the ARN framework.

\section{AR-mediated Negotiation-based Framework}

Multi-agent systems require the agents, including humans and robots, to extensively communicate with each other, because of the inter-dependency among the agents' actions. 
The inter-dependency can be in the form of state constraints, action synergies, or both.
In this section, we describe our augmented reality-mediated, negotiation-based  (ARN) framework, as shown in Figure~\ref{fig:framework}, that utilizes the AR interface (Section~\ref{sec:ar_interface}) capabilities to enable multi-turn, bidirectional communication between human and robot teammates, and iterative “negotiation” toward the most effective human-robot collaboration behaviors.

\vspace{.8em}
\noindent{\bf{The ARN Algorithm: }}
ARN is our general solution that generates plans for all the robots while taking human feedback into account as constraints.
The input of Algorithm~\ref{alg:arn_algorithm} includes the initial state of $N$ robots ($s$), a set of goal states ($\boldsymbol{G}$), Task Planner $MRP^{T}$, Motion Planner $MP$, and, $R$ which is a set of $N$ robot teammates.

In Lines~\ref{line:initialize_C}-\ref{line:initialize_trj}, ARN first initializes three empty lists, $C$, $\boldsymbol{Pose}$, and $\boldsymbol{Trj}$.
Then in Line~\ref{line:generate_initial_plans}, ARN generates a set of initial plans ($\boldsymbol{P}$) for the team of $N$ robots using the task planner ($MRP^{T}$).
Entering a while-loop in Line~\ref{line:main_while_loop}, ARN runs until all robots have completed their tasks.
Next, in Line~\ref{line:for_loop_robots}, ARN enters a for-loop with $N$ iterations.
In Line~\ref{line:goal_reached}, ARN initially checks if the $i$th robot has reached the goal or not.
If not, Lines~\ref{line:see_front_action}-\ref{line:end_if_action_complete} are executed, where ARN first gets the current action ($a_{i}$) of the $i$th robot using $\boldsymbol{P}$.
$MP$ takes $a_{i}$ as input and generates a motion trajectory, which is then added to $\boldsymbol{Trj}$.
In Line~\ref{line:get_pose_robots}, ARN gets the pose of the $i$th robot using the $\boldsymbol{getPose}$ function.
The poses of all the robots are stored in $\boldsymbol{Pose}$.
The if-statement in Line~\ref{line:pop_action} checks if the current action is complete or not, and if it is complete, then the action is popped from the plan queue.

In Line~\ref{line:Visualization_Agent}, ARN passes the generated set of trajectories and poses to the $\boldsymbol{Vslz}$ function (Section~\ref{sec:vslz}), which then renders a frame ($V$) on the AR device to augment the motion plans and live locations.
Then using $\boldsymbol{getFdk}$ function, ARN checks if any human feedback is available and stores it in $H$(Line~\ref{line:getHumanFeedback}).
Based on $H$, a new set of activated constraints $C^{'}$ are obtained using the $\boldsymbol{Cnsts}$ function (Section~\ref{sec:cnsts}).
If the new set of activated constraints are different from the old activated constraints, then ARN generates new plans for the team of robots using the task planner (Line~\ref{line:generateNewPlans}), and then the old set of constraints are replaced with the new ones (Line~\ref{line:update_constrained_resources}).
Finally, in Line~\ref{line:all_robot_goals}, ARN checks if all the robots have reached their goals.
If the robot(s) have not reached their goals, ARN continues to the next iteration of the while loop.
Otherwise, the \emph{break} statement exits the while loop, marking the end of the algorithm (Line~\ref{line:break_statement}).

Algorithm~\ref{alg:arn_algorithm} enables the bidirectional communication within human-robot teams toward effective human-robot collaboration.
Two important steps in ARN include Lines~\ref{line:generate_trajectory} and \ref{line:generateNewPlans} for motion planning and task planning respectively.

\begin{algorithm}[h] \footnotesize
\caption{ARN Framework}\label{alg:arn_algorithm}

\textbf{Input}: $s$, $\boldsymbol{G}$, $MRP^{T}, MP$, $R$ 
\begin{algorithmic}[1]
\State {Initialize an empty list of activated constraints $C$ as $\emptyset$} \label{line:initialize_C}
\State {Initialize an empty list of $\boldsymbol{Pose}$ to store current pose of all robots.}\label{line:initialize_pose}
\State{Initialize an empty list of size $N$: $\boldsymbol{Trj}$, where $Trj \in \boldsymbol{Trj}$}\label{line:initialize_trj}

\State{$\boldsymbol{P} = MRP^{T}(s, \boldsymbol{G}, C)$, where $p \in \boldsymbol{P}$, and $p$ is a queue to store the action sequence of every robot } \label{line:generate_initial_plans}


\While{$True$} \label{line:main_while_loop}
\For{{\bf{each}} $i \in [0, 1, \cdots, N$-$1]$}\label{line:for_loop_robots}
\If{$\boldsymbol{goalReached}(s_{i}, G) == False$} \label{line:goal_reached}
\State $a_{i} \xleftarrow{} \boldsymbol{P}[i].front()$ \label{line:see_front_action}
\State $R[i]$ executes $a_{i}$
\State $\boldsymbol{Trj}[i] \xleftarrow{} MP(a_{i})$  \label{line:generate_trajectory} 
\State $\boldsymbol{Pose}[i] \xleftarrow{} \boldsymbol{getPose}(R[i])$\label{line:get_pose_robots}
\If{$a_{i}$ is completely executed}
\State{$\boldsymbol{P}[i].pop()$}\label{line:pop_action}
\EndIf \label{line:end_if_action_complete}

\EndIf

\EndFor
\State{$V \xleftarrow{} \boldsymbol{Vslz}(\boldsymbol{Trj, Pose})$} \label{line:Visualization_Agent} \algorithmiccomment{V is a frame rendered on AR device}
\State{$H \xleftarrow{} \boldsymbol{getFdk}()$}\label{line:getHumanFeedback} \algorithmiccomment{Get feedback from AR device}

\State{$C^{'} \xleftarrow{} \boldsymbol{Cnsts}(H)$} \label{line:get_constraints}

\If{$C^{'} \neq C$}
    \State{$\boldsymbol{P} \xleftarrow{} MRP^{T}(s, \boldsymbol{G}, C)$}\label{line:generateNewPlans} \algorithmiccomment{Update Plans}
    \State{$C \xleftarrow{} C^{'}$}\label{line:update_constrained_resources}
\EndIf

\If{$s_{i}$ for all robots $\in G$} \algorithmiccomment{Check if all robots reached goals} \label{line:all_robot_goals}
\State{break}\label{line:break_statement}
\Else{}
\State{continue}
\EndIf

\EndWhile

\end{algorithmic}
\end{algorithm}

\begin{figure*}[t]
\vspace{.4em}
    \begin{center}
    \subfigure[]
    {\includegraphics[height=3.4cm]{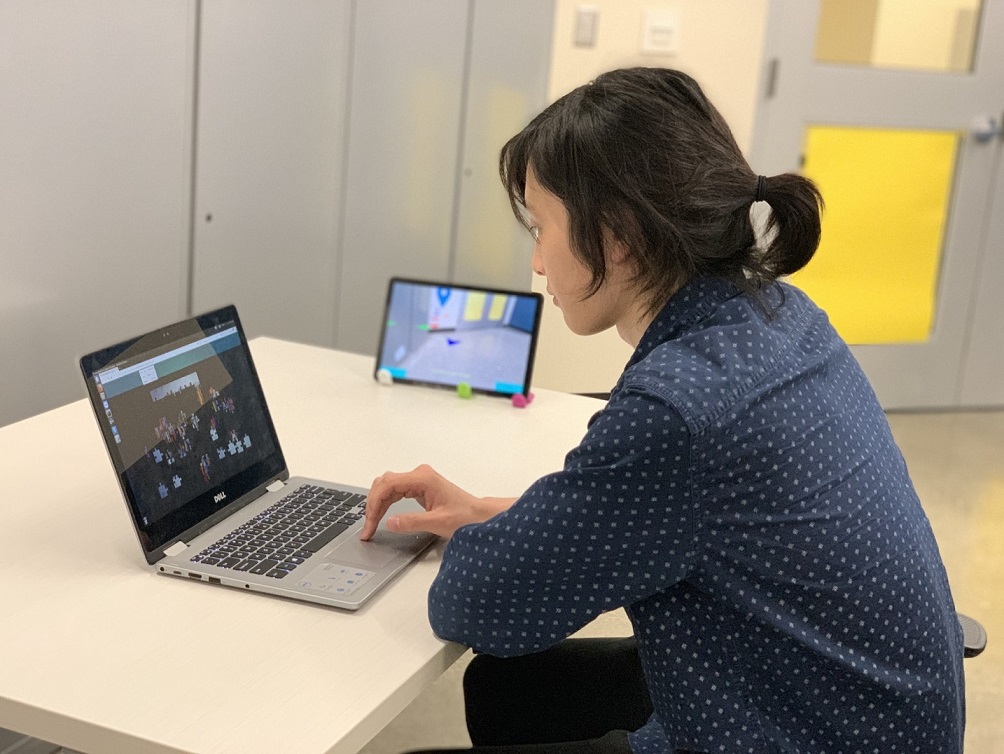}
    \label{fig:participant_solving_jigsaw}}
    \subfigure[]
    {\includegraphics[height=3.4cm]{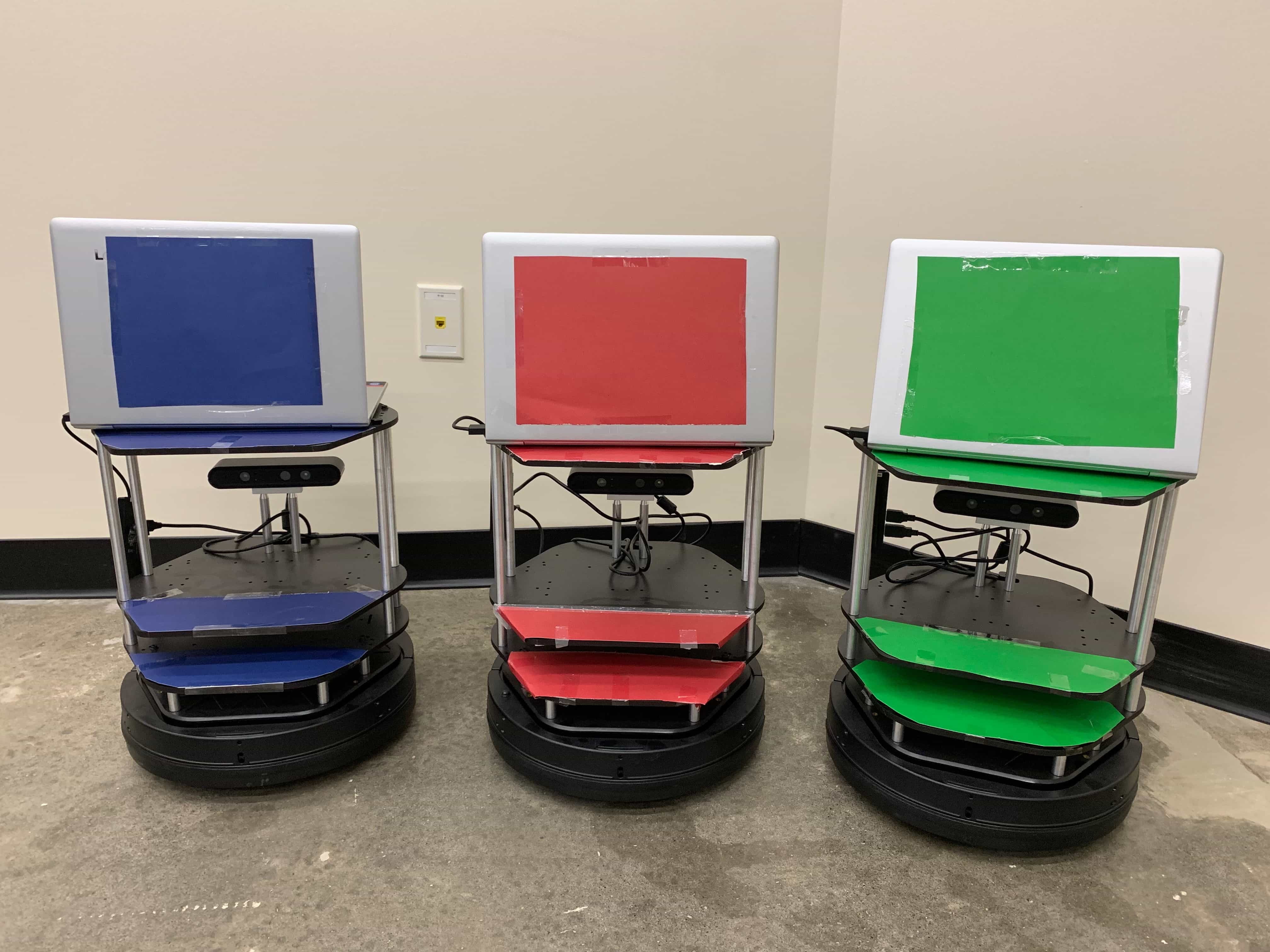}
    \label{fig:turtlebots}}
    \subfigure[]
    {\includegraphics[height=3.4cm]{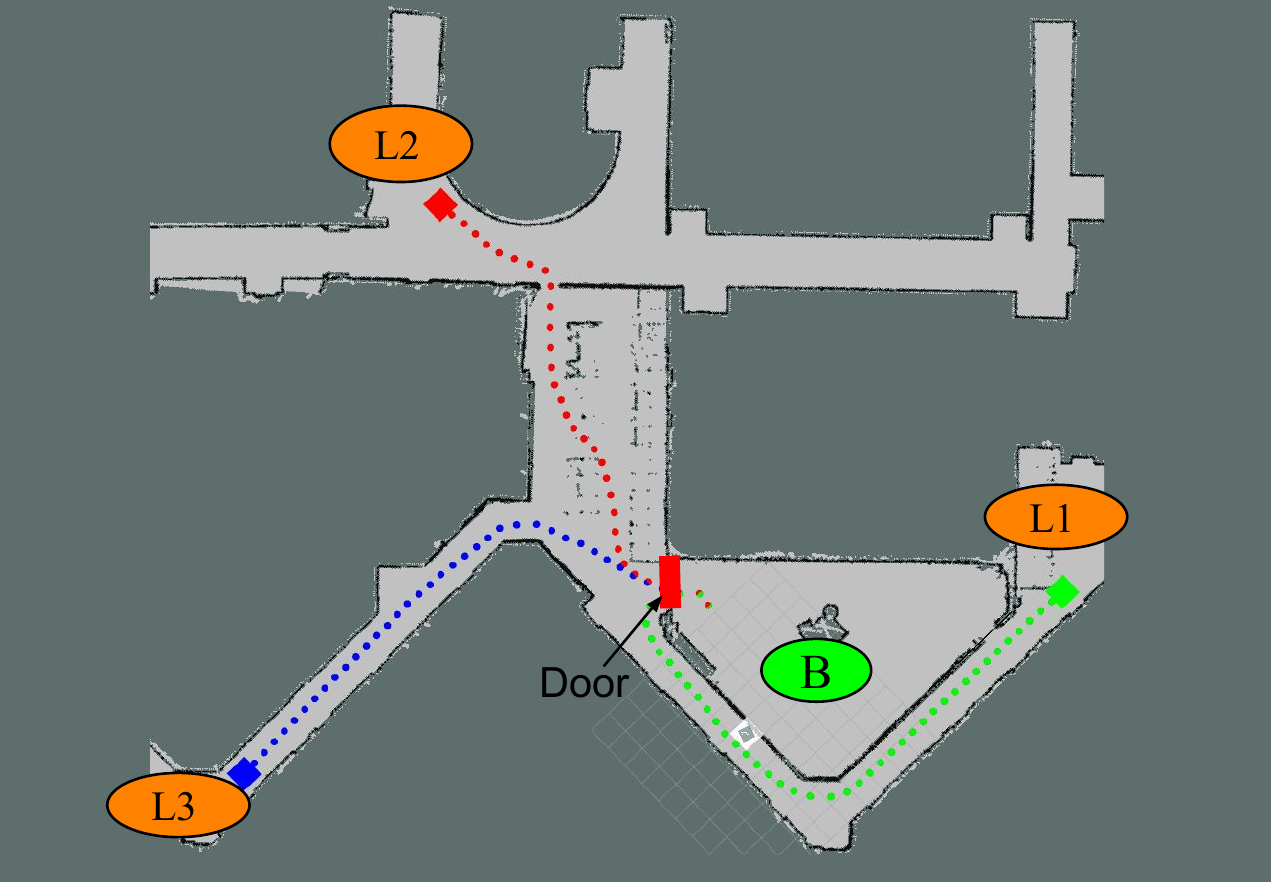}
    \label{fig:map}}
    \vspace{-.8em}
    \caption{(a) A human playing the Jigsaw game alongside our AR interface running on a tablet, where the door with yellow paper on it blocks the robots from entering the room by themselves; 
    (b) Turtlebot-2 robot platforms used in the real-world experiments; and 
    (c) Our domain map, including the three ``loading'' locations ($L1$, $L2$, and $L3$), and base station $B$, where the human participants work on Jigsaw games in the ``base station'' room and help open the door (in red color) as they feel necessary. The colored trajectories are presented to participants in the non-AR baseline approach.}
    
    \label{fig:setup}
    \end{center}
    \vspace{-1.5em}
\end{figure*}


\vspace{.8em}
\noindent{\bf{Multi-robot Task Planning: }}
Multi-robot task planning ($MRP^{T}$) is for computing plans for a team of $N$ robots.
The input of $MRP^{T}$ includes the initial state of $N$ robots ($s$), a set of goal states ($\boldsymbol{G}$), and a set of activated constraints ($C$).
$MRP^{T}$ returns a set of plans $\boldsymbol{P}$, where $p \in \boldsymbol{P}$ consists of a set of actions $[a_{1}, a_{2}, \cdots, a_{n}]$, and $a_i$ is the action for the $i$th robot. 
Note that $p$ is a queue used for storing a robot's action sequence. 
$MRP^{T}$ can be separately developed, and its development is independent of ARN. 
Our implementation of $MRP^{T}$ is introduced in Section~\ref{sec:ins}. 

\vspace{.8em}
\noindent{\bf{Motion Planning: }}
To visualize the motion plans of the robots, we need to convert the robot actions obtained from $MRP^{T}$ to a set of trajectories that can be visualized in the AR device.
For this purpose, we use a motion planner $MP$, which takes an action $a_{i}$ as input and returns the trajectory that the robot needs to follow to complete the action.
$$
\boldsymbol{Trj} \leftarrow MP(a), \textnormal{~where~} Trj=[L_1, L_2, \cdots, L_M]
$$

The trajectory is a set of locations, and every location $L_m$ is a set of $x$ and $y$ coordinates for representing a sequence of $M$ 2D coordinates. 
The $\boldsymbol{getPose}(R[i])$ function returns the current pose of the $i$th robot.
The poses of the team of $N$ robots are stored in $\boldsymbol{Pose}$, where $\boldsymbol{Pose}[i]$ represents the pose of the $i$th robot.

In this section, we have described our main contribution ARN for HRC, and how each of the above three functions is used in ARN.
But, ARN can be implemented in many ways, and now in the following section, we describe how we implemented ARN in our domain.

\section{Multi-Robot Task Planning}
\label{sec:ins}

\noindent\textbf{Single Robot Task Planning:}
We use Answer Set Programming (ASP) for robot task planning.
ASP is a popular declarative language for knowledge representation and reasoning (KRR)~\cite{gelfond2014knowledge,lifschitz2008answer}, and has been applied to a variety of planning problems~\cite{yang2014planning, lifschitz2002answer,erdem2016applications}, including robotics~\cite{erdem2018applications}.
ASP is particularly useful for robot task planning in domains that include a large number of objects~\cite{jiang2019task}. 
We formulate five actions in our domain: \texttt{approach}, \texttt{opendoor}, \texttt{gothrough}, \texttt{load}, and \texttt{unload}.
For instance, 

    \vspace{.5em}
    \texttt{open(D,I+1):$-$ opendoor(D,I).}
    \vspace{.5em}

\noindent
states that executing action \texttt{opendoor(D,I)} causes door \texttt{D} to be open at the next step. 
The following states that a robot cannot execute the \texttt{opendoor(D)} action, if it is not facing door \texttt{D} at step \texttt{I}. 

\vspace{.5em}
\texttt{:$-$ opendoor(D,I), not facing(D,I).}
\vspace{.5em}


The following shows an example plan for the robot generated by the ASP planner: 
\texttt{approach(D,0). opendoor(D,1). gothrough(D,2).}, 
which suggests the robot to first approach door \texttt{D}, then open door \texttt{D}, and finally go through the door at step \texttt{I}=\texttt{2}. 

\vspace{.8em}
\noindent\textbf{Multi-Robot Task Planning:}
Building on this single-robot task planner, we use an algorithm called IIDP (\emph{iterative inter-dependent planning}) to compute joint plans for robot teams~\cite{jiang2019multi}, where IIDP is a very efficient (though sub-optimal) multi-robot planning algorithm. 
Details are omitted from this paper due to the page limit. 

It should be noted that the main contribution of this research is on the AR interface and the ARN framework, and this section is only for completeness. 

\begin{figure*}[t]
\vspace{.4em}
\begin{center}
    \subfigure[]
    {
    \hspace*{-3em}
    \includegraphics[height=3.4cm]{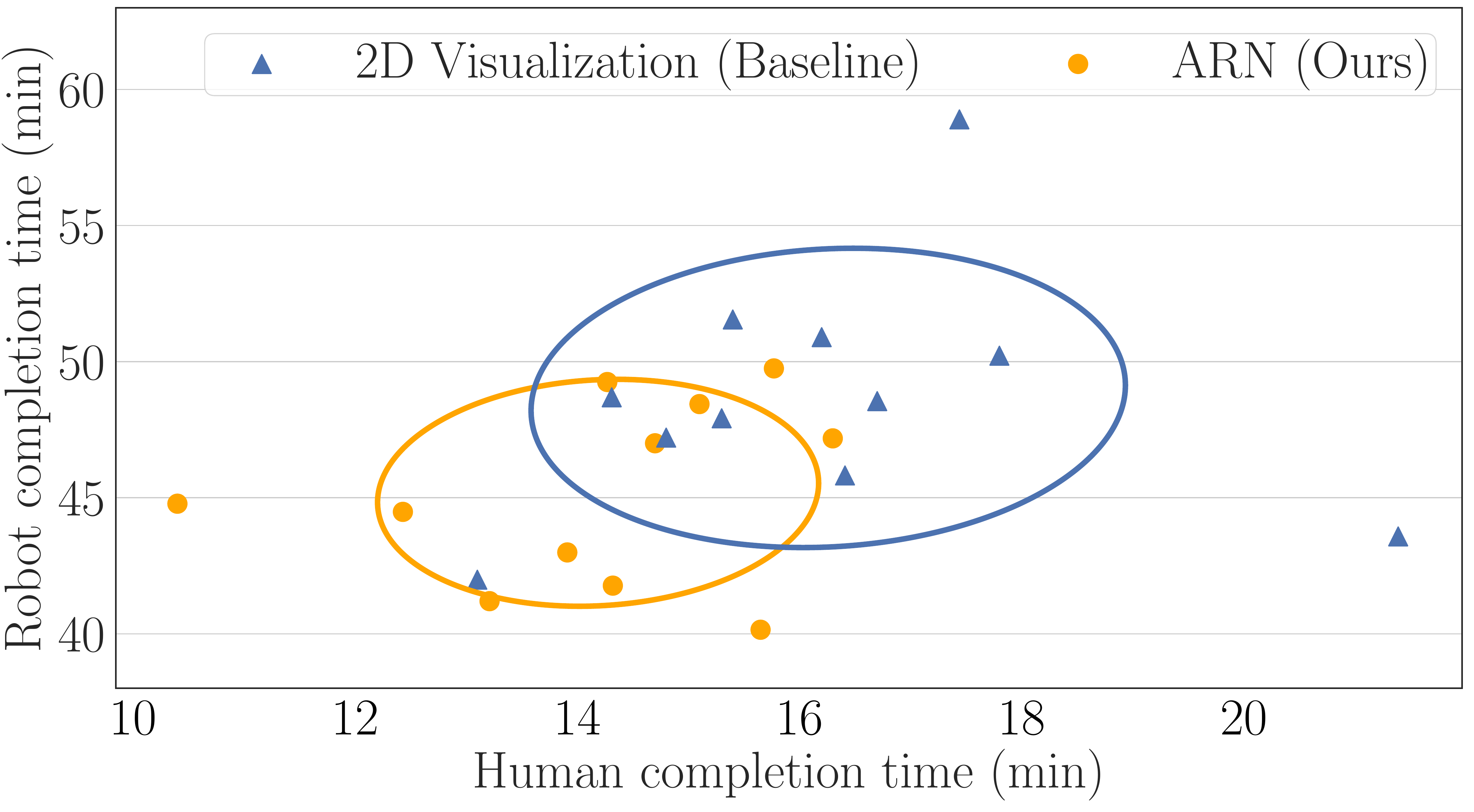}
    \label{fig:scatter_plot}}
    \hspace{10mm}
    \subfigure[]
    {\includegraphics[height=3.4cm]{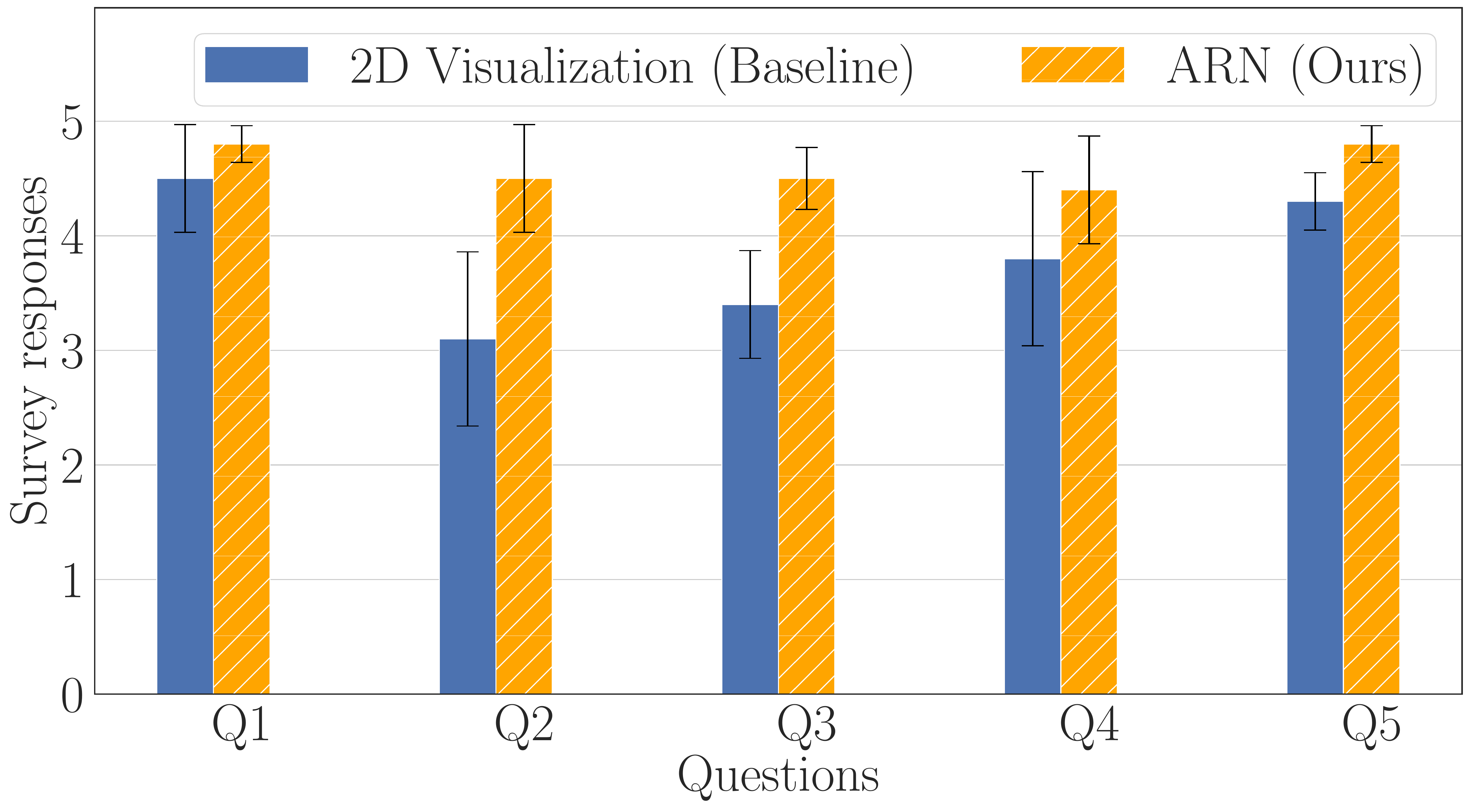}
    \label{fig:bar_graph}}
    \vspace{-.8em}
    \caption{(a) Overall performance in task completion time using ARN and a 2D visualization baseline interface (Rviz-based) in human participant study; and
    (b)~Survey responses from participants collaborating with a team of robots on delivery tasks.
    }
    \label{fig:results}
    \end{center}
    \vspace{-1.5em}
\end{figure*}

\section{Experiment Setup and Results}

Experiments have been conducted in simulation and on real robots to evaluate the following two hypotheses in comparison to non-AR baselines: I) ARN improves the overall efficiency in human-robot collaboration, and II) ARN produces better experience for non-professional users. 


\subsection{Real-world Experiments}

\noindent\textbf{Experiment Setup:}
Fig.~\ref{fig:participant_solving_jigsaw} shows a human user playing the ``Jigsaw'' game on a computer, while three Turtlebot-2 robots work on delivering three objects from three different locations to a base station, as shown in Fig.~\ref{fig:turtlebots}.
The map of this shared office environment is shown in Fig.~\ref{fig:map}. 
This delivery task requires human-robot collaboration, because both human and robots have their own non-transferable tasks, and that the robots need people to help open a door to ``unload'' objects to the station. 
Since the robots do not have an arm, they simply visit places, and loading and unloading actions were not conducted in the real world. 
All software runs on Robot Operating System (ROS)~\cite{quigley2009ros}, while door-related functionalities were built on the BWI code base~\cite{khandelwal2017bwibots}. 


Eleven participants of ages 20-30 volunteered to participate in the experiment, including four females and seven males.\footnote{The experiments were approved by the Institutional Review Board (IRB).}
The participants worked in collaboration with the robots to \emph{minimize each individual team member's task-completion time}.
It should be noted that this setting covers the goal of minimizing the slowest team member's task completion time, but goes beyond that to capture efficient team members' task completion times.

Each participant used both a baseline system (to be described), and our ARN system to work on the collaborative delivery tasks, where the two systems are randomly ordered to avoid discrepancies in the results caused by people's learning curves. 
None of the participants have any experience of working with the robots. 
The Jigsaw puzzles were generated randomly, and no puzzle was repeated throughout the experiment. 
The settings ensure a fair comparison between the baseline approach and ARN on human-robot collaboration.

\vspace{.8em}
\noindent{\bf{Baseline (2D Visualization):}}
Without the AR interface, one can use a standard 2D visualization interface to track the status of the robot team. 
Our non-AR baseline system was built on the visualization tool of Rviz that has been widely used within the ROS community~\cite{rvizROSW20:online}.
The participants could see the robots' planned trajectories along with their locations in a 2D map on a laptop, as shown in Fig.~\ref{fig:map}.
According to the robots' locations and planned trajectories, the participants decide when to let the robots in through door-opening actions.
It should be noted that the participants can potentially find it difficult to map the robots' locations shown on the map to real-world locations. 
In comparison, the AR interface very well supports the communication of spatial information. 
For instance, it is highly intuitive to derive the robot's real locations from Fig.~\ref{fig:visualizer_figure}. 


\vspace{.8em}
\noindent{\bf{Results on Overall Task Completion Time:}}
Fig.~\ref{fig:scatter_plot} shows the overall performance of ARN compared to the baseline.
The x-axis corresponds to the participants' average task completion time, and the y-axis corresponds to the sum of the three robots' task completion time in total, i.e., $T^{R1} + T^{R2} + T^{R3}$.
Each data point corresponds to one trial (using ARN or baseline). 
The two ellipses show their 2D standard deviations. 
We see that the data points of ARN are close to the bottom-left corner, which supports Hypothesis-I (ARN improves the overall collaboration efficiency). 

Looking into the results reported in Fig.~\ref{fig:scatter_plot}, we calculated the total time of each trial, i.e., 
$$
    T^{all} = T^H + T^{R1} + T^{R2} + T^{R3}
$$
where $T^H$ and $T^{Ri}$ are the task completion times of the human and the $i$th robot respectively. 
ARN significantly improved the performance in $T^{all}$ in comparison to the 2D-visualization baseline, where $p$-value is $.011$. 
This shows that ARN performs \emph{significantly} better than the baseline in total task completion time.

\vspace{.8em}
\noindent{\bf{Questionnaires:}}
At the end of each experiment trial, participants were asked to fill out a survey form indicating their qualitative opinion over the following items. 
The response choices were: 1 (Strongly disagree), 2 (Somewhat disagree), 3 (Neutral), 4 (Somewhat agree), and 5 (Strongly agree).

The questions include: 
Q1, \emph{The tasks were easy to understand}; 
Q2, \emph{It was easy to keep track of robot status}; 
Q3, \emph{I could focus on my task with minimal distraction from robot}; 
Q4, \emph{The task was not mentally demanding (e.g., remembering, deciding, thinking, etc.)}; and 
Q5, \emph{I enjoyed working with the robot and would like to use such a system in the future.} 
It should be noted that Q1 is a question aiming at confirming if the participants understood the tasks, and is not directly relevant to the evaluation of our hypotheses.

\begin{figure*}[t]
\vspace{.4em}
\begin{center}
    \subfigure[]
    {\includegraphics[height=2.4cm]{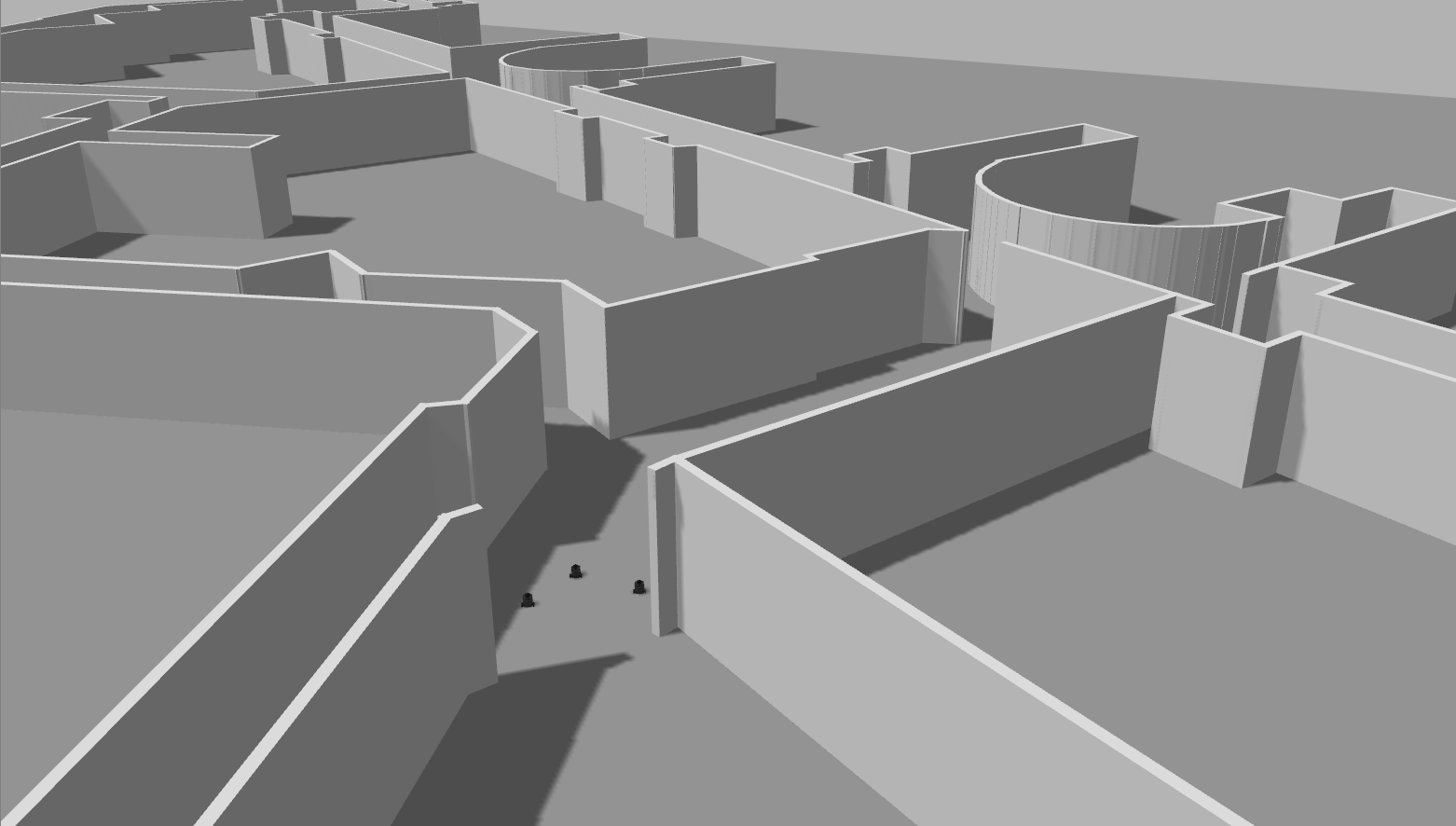}
    \label{fig:gazebo_robots}}
    \subfigure[]
    {\includegraphics[height=2.4cm]{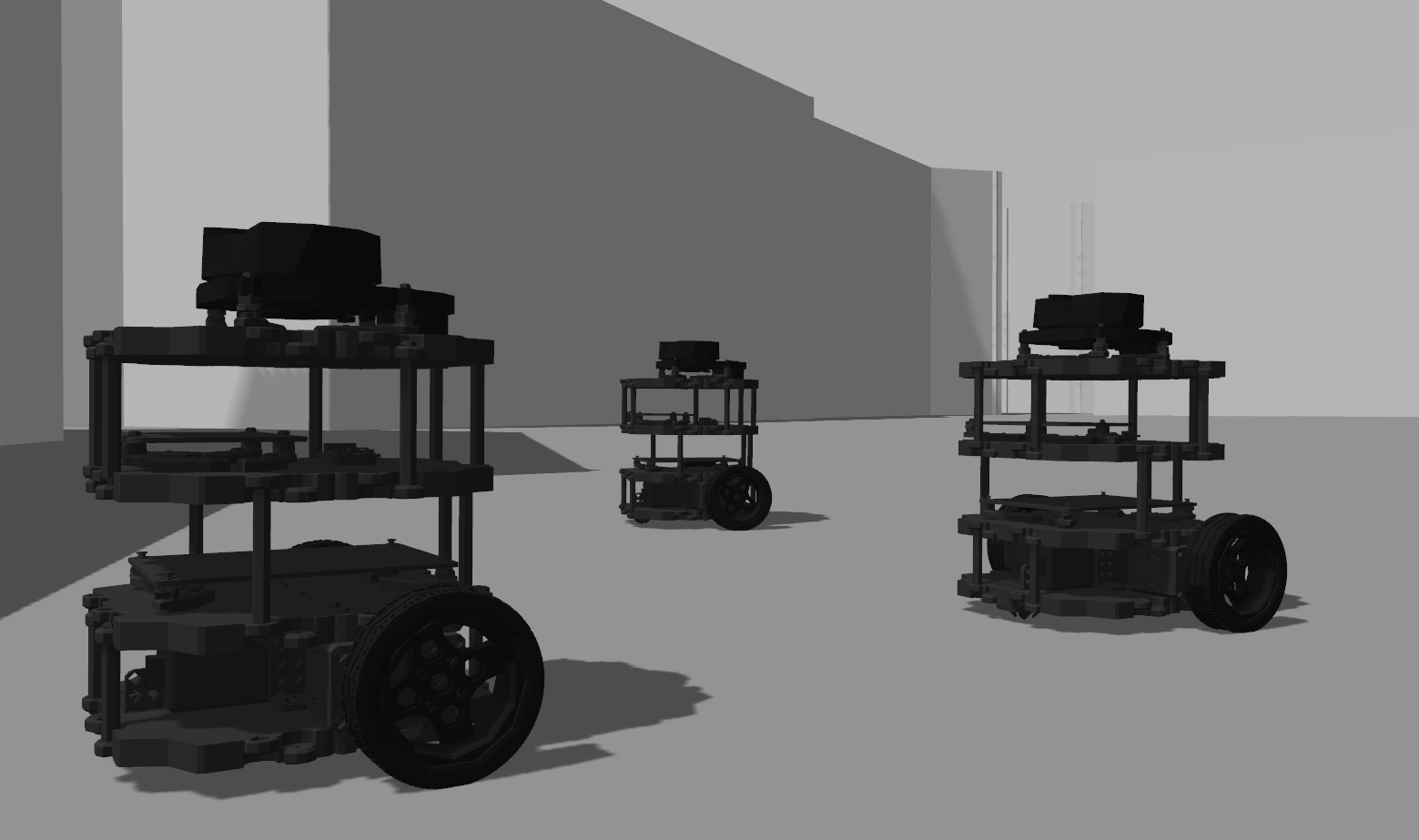}
    \label{fig:zoomed_robots}}
    \subfigure[]
    {\includegraphics[height=2.4cm]{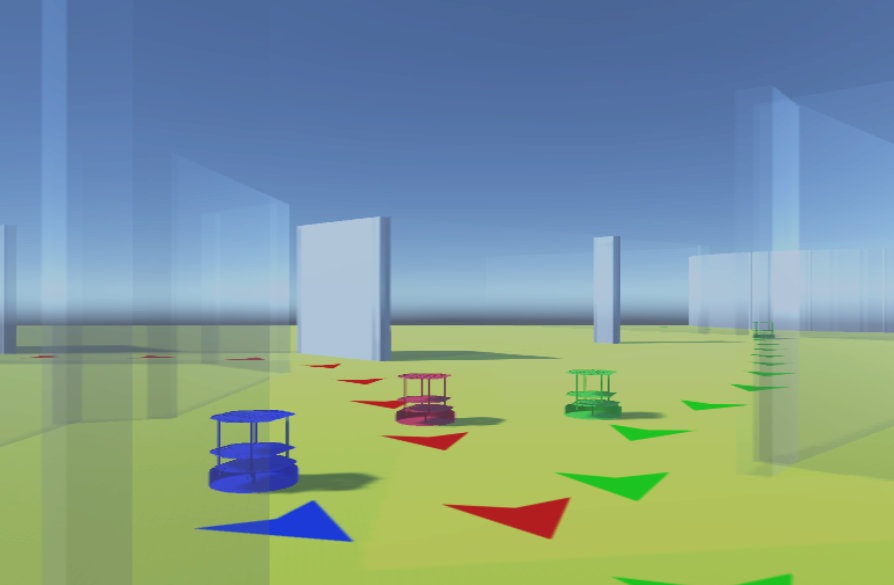}
    \label{fig:enlarged_virt_tab}}
    \subfigure[]
    {\includegraphics[height=2.4cm]{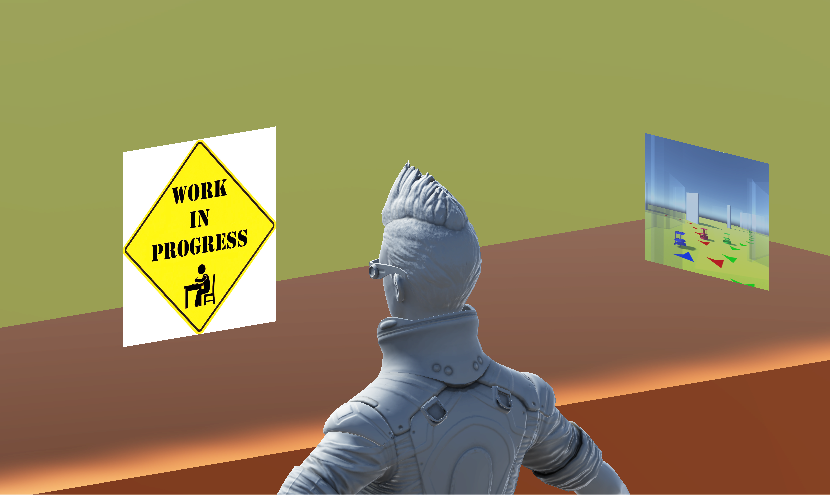}
    \label{fig:unity_world_with_table_doors}}
    \vspace{-.8em}
    \caption{(a) Gazebo showing the three robots waiting outside the base station; 
    (b) Enlarged view of the three robots in Gazebo; 
    (c)~Enlarged view of virtual AR device showing the three robots waiting outside the base station; and 
    (d)~Unity world with the virtual human, and the AR device placed on the table.
    }
    \label{fig:sim_figs}
    \end{center}
    \vspace{-1.5em}
\end{figure*}

Fig.~\ref{fig:bar_graph} shows the average scores from the human participant surveys.
Results show that ARN produced higher scores on Questions Q2-Q5. 
We also calculated the corresponding $p$-values and observed significant improvements in Q2, Q3, and Q5 with the $p$-values $< .001$.
The significant improvements suggest that ARN helps the participants keep track of the robot status, is less distracting, and is more user-friendly to non-professional participants. 
However, the improvement in Q4 was not significant, and one possible reason is that making quantitative comparisons over the ``mentally demanding'' level can be difficult for the participants due to the two interfaces being very different by nature.\footnote{This submission includes a supplementary demo video for illustrating a complete human-robot collaboration trial.}

\subsection{Simulation Experiments}

\noindent{\bf{Experiment Setup:}}
Experiments with human participants have shown the effectiveness of ARN. 
We are interested in how much the ``negotiation-based'' component contributes to the success of ARN. 
Considering the practical difficulties in large-scale experiments with human-multi-robot systems, we have built a simulation platform for extensive evaluations. 
In the simulation, we compared two configurations of ARN: one activated the ``Restrictor'' for adding constraints from human feedback, and the other did not allow the robots to take human feedback.
The two configurations are referred to as ``w/ fdbk'' and ``w/o fdbk'' respectively. 
Each configuration was evaluated with one hundred trials. 

\vspace{.8em}
\noindent{\bf{Simulation of Robot Behaviors:}}
Fig.~\ref{fig:gazebo_robots} and Fig.~\ref{fig:zoomed_robots} present the environment used for simulating robots' navigation behaviors, where we use Gazebo to build the simulation environment~\cite{koenig2004design}. 
Our simulation environment exactly replicates the real-world environment, where the robots work on the delivery tasks (Fig.~\ref{fig:map}).
We use an implementation of the Monte Carlo Localization algorithm, the \emph{amcl}~\cite{amclROSW92:online} package of ROS for robot localization~\cite{dellaert1999monte}, and \emph{move base}~\cite{movebase65:online} package of the ROS navigation stack for robot path planning.

\vspace{.8em}
\noindent{\bf{Simulation of AR Interface:}}
Fig.~\ref{fig:enlarged_virt_tab} shows the augmented robots waiting at the door in the virtual AR device.
The AR interface is simulated using Unity~\cite{UnityRea19:online}.
The virtual AR device is placed on the table and can be rotated left or right, similar to the real-world setup. 
The virtual AR device supports ARN features, such as visually augmenting the ``real world'' with the robots' current locations and the planned motion trajectories, and taking human feedback and converting feedback into task constraints.

\vspace{.8em}
\noindent{\bf{Simulation of Human Behaviors:}}
Human behaviors are simulated, as shown in Fig.~\ref{fig:unity_world_with_table_doors}.
In each trial, the virtual human is engaged in its own dummy task, while it also has to help the robots to enter the \emph{base station} by opening the door.
The door in the simulation environment is not visualizable, but we simulate the door opening actions.
The virtual human can take one action at a time from the following six actions: A1, \emph{Work on his/her own task (simplified as staring at the poster in Unity)}; 
A2, \emph{Tilt the AR device to the left}; 
A3, \emph{Tilt the AR device to the right};
A4, \emph{Give feedback ``I will be busy for two minutes''};
A5, \emph{Give feedback ``I will be busy for four minutes''}; and
A6. \emph{Open the door}. 
In the ``without feedback'' configuration, the virtual human's action space does not include the two feedback actions (A4 and A5).

Prior to every trial, we sample $T^H$ (human's total task completion time) from a Gaussian distribution, where the mean and variance were empirically selected. 
While the virtual human is ``working'' on his/her own task, we sample the time of clicking one of ``feedback'' buttons (A4 or A5) from another Gaussian distribution, where A4 and A5 are randomly selected. 
The door opening action can be independently triggered by each individual robot that is waiting outside the door in probability $0.6$, where we sample the door-opening action (to open it or not) every $20$ seconds. 
Due to the independency, the human is more likely to open the door when there are more robots waiting outside. 

After the ``human'' takes action A4 (``busy for two minutes), the probability of the human opening the door for each robot is significantly reduced to only $0.2$, and this probability is changed back to $0.6$ after the ``two minutes'' busy time. 
Finally, when the human is done with his/her own task, this door-opening probability is increased to $0.9$.

\begin{table}[tbh] \footnotesize
    \caption{Simulation Results}
    \vspace{-.5em}
    \centering
    \begin{tabular}{|l|c|c|c|} 
    \hline
    Method&$T^{H}$&$T^{all}$&$T^{R_{last}}$ \\
    \hline
         ARN (w/ feedback) & 15.67 (3.04) & 46.64 (4.31) & 16.65 (0.73)\\
     \hline
         ARN (w/o feedback) & 16.77 (4.73) & 50.43 (6.35)& 17.94 (0.90)\\  
     \hline
    \end{tabular}
     \vspace{-1em}
    \label{tab:sim_results}
\end{table}

\vspace{.8em}
\noindent{\bf{Results:}}
Table~\ref{tab:sim_results} shows the average task completion time of the virtual human ($T^{H}$), the average task completion time of the team of robots ($T^{all}$), and the average task completion time of the last robot ($T^{R_{last}}$). 
The two rows correspond to the two configurations: ARN with feedback, (i.e., Restrictor being activated), and ARN without feedback (i.e., Restrictor not being activated). 
The average completion times are lower for all the trials of ARN with feedback, as compared to the ``without feedback'' configuration.
Furthermore, we calculated the total time of each trial, i.e., $T^{all} = T^H + T^{R1} + T^{R2} + T^{R3}$.
We observed significant improvements in the overall task completion time ($T^{all}$) with a $p$-value $<.001$ in the comparisons between the two ARN configurations. 
This shows that the feedback feature of ARN significantly contributes to the effectiveness of ARN in multi-turn negotiation-based collaboration behaviors.

\section{Conclusions}
In this paper, we introduce a novel augmented reality-mediated, negotiation-based framework, called ARN, for human-robot collaboration tasks. 
The human and robot teammates work on non-transferable tasks, while the robots have limited capabilities and need human help at certain phases for task completion. 
ARN enables human-robot negotiations through visualizing robots' current and planned actions, while also incorporating the human feedback into robot re-planning. 
Experiments in simulation and with human participants show that ARN significantly increased the overall efficiency of human-robot collaboration, in comparison to a non-AR approach from the literature. 


\bibliographystyle{IEEEtran}
\bibliography{ref}

\end{document}